\PassOptionsToPackage{table, dvipsnames}{xcolor}
\documentclass[sigconf,nonacm=true,natbib=true,pbalance=true]{acmart}

\usepackage{amsmath}
\usepackage{physics}
\usepackage{dsfont}
\usepackage{mathrsfs}
\usepackage{wasysym}
\usepackage{adjustbox}

\usepackage{lineno,hyperref}
\usepackage{float}
\usepackage{multirow}
\usepackage{xcolor,colortbl}
\usepackage{adjustbox}
\usepackage{booktabs}

\usepackage{graphicx,lipsum,afterpage,subcaption}
\usepackage{enumitem}
\usepackage{xspace}
\usepackage{float}
\usepackage{tabularx} 

\expandafter\let\csname c@tblerows\endcsname\rownum

\usepackage{makecell}
\usepackage{boldline}
\usepackage{cleveref}
\usepackage{stackengine}

\usepackage[flushleft]{threeparttable}

\usepackage{bm}
\usepackage{multirow}
\usepackage{graphicx}

\usepackage{algorithm}
\usepackage[noend]{algpseudocode}

\usepackage[htt]{hyphenat}

\usepackage[frozencache=false,cachedir=.,newfloat]{minted}

\usepackage[table]{xcolor}

\usepackage[subtle]{savetrees}

\definecolor{White}{gray}{0.995} \usepackage{colortbl}

\usepackage[font=small,skip=1pt]{caption}
\usepackage{listings}
\usepackage{pifont} 

\usepackage{chato-notes} 

\usepackage[T1]{fontenc} 
\SetupFloatingEnvironment{listing}{name=Configuration}
\usepackage{csquotes}

\usepackage{tcolorbox} 
\usepackage{xcolor}    
\usepackage{soul}
\usepackage{balance}

\newcommand{\cmark}{\ding{51}}

\newcommand{\framework}{\textbf{WarpRec}}

\definecolor{LightGray}{gray}{0.9}
\definecolor{OOMColor}{RGB}{255, 230, 230} 
\definecolor{applegreen}{rgb}{0.55, 0.71, 0.0}
\definecolor{JungleGreen}{rgb}{0.16, 0.57, 0.5}
\definecolor{Thistle}{rgb}{0.85, 0.75, 0.85}

\providecommand{\myrowcolour}{\rowcolor{gray!15}}

\AtBeginDocument{%
  }

\iffalse 
\setcopyright{acmlicensed}
\else
\setcopyright{none}
\fi
\copyrightyear{2026}
\acmYear{2026}
\acmDOI{XXXXXXX.XXXXXXX}
\acmConference[SIGIR '26]{SIGIR}{July 20--24, 2026}{Melbourne, AU}
\acmISBN{978-1-4503-XXXX-X/2018/06}

\begin{document}

\title{WarpRec: Unifying Academic Rigor and Industrial Scale for Responsible, Reproducible, and Efficient Recommendation}


\settopmatter{authorsperrow=4}

\author{Marco Avolio}
\affiliation{%
  \institution{Wideverse, Italy}
  \country{}
}
\authornote{These 
authors contributed equally to this research.}
\orcid{0009-0001-0046-6746}

\author{Potito Aghilar}
\authornotemark[1]
\affiliation{%
  \institution{Politecnico di Bari, Italy}
  \country{}
}
\affiliation{%
  \institution{Wideverse, Italy}
  \country{}
}
\orcid{0000-0003-3662-0802}

\author{Sabino Roccotelli}
\authornotemark[1]
\affiliation{%
  \institution{Politecnico di Bari, Italy}
  \country{}
}
\affiliation{%
  \institution{Wideverse, Italy}
  \country{}
}
\orcid{0009-0003-3479-9607}

\author{Vito Walter Anelli}
\authornotemark[1]
\affiliation{%
  \institution{Politecnico di Bari, Italy}
  \country{}
}
\orcid{0000-0002-5567-4307}

\author{Chiara Mallamaci}
\affiliation{%
  \institution{Politecnico di Bari, Italy}
  \country{}
}
\orcid{0009-0003-2663-3068}

\author{Vincenzo Paparella}
\affiliation{%
  \institution{ISTI-CNR, Pisa, Italy}
  \country{}
}
\orcid{0000-0002-5602-1682}

\author{Marco Valentini}
\affiliation{%
  \institution{Politecnico di Bari, Italy}
  \country{}
}
\orcid{0009-0004-1652-0745}

\author{Alejandro Bellogín}
\affiliation{%
  \institution{UAM, Madrid, Spain}
  \country{}
}
\orcid{0000-0001-6368-2510}

\author{Michelantonio Trizio}
\affiliation{%
  \institution{Wideverse, Italy}
  \country{}
}
\orcid{0009-0005-5363-2647}

\author{Joseph Trotta}
\affiliation{%
  \institution{OVS, Italy}
  \country{}
}
\orcid{0009-0004-2731-6926}

\author{Antonio Ferrara}
\affiliation{%
  \institution{Politecnico di Bari, Italy}
  \country{}
}
\orcid{0000-0002-1921-8304}

\author{Tommaso Di Noia}
\affiliation{%
  \institution{Politecnico di Bari, Italy}
  \country{}
}
\orcid{0000-0002-0939-5462}

\renewcommand{\shortauthors}{Avolio et al.}

\begin{abstract}
Innovation in Recommender Systems is currently impeded by a fractured ecosystem, where researchers must choose between the ease of in-memory experimentation and the costly, complex rewriting required for distributed industrial engines.
To bridge this gap, we present \textsc{WarpRec}, a high-performance framework that eliminates this trade-off through a novel, backend-agnostic architecture. It includes $50+$ state-of-the-art algorithms, $40$ metrics, and $19$ filtering and splitting strategies that seamlessly transition from local execution to distributed training and optimization. The framework enforces ecological responsibility by integrating CodeCarbon for real-time energy tracking, showing that scalability need not come at the cost of scientific integrity or sustainability.
Furthermore, \textsc{WarpRec} anticipates the shift toward Agentic AI, leading Recommender Systems to evolve from static ranking engines into interactive tools within the Generative AI ecosystem. 
In summary, \textsc{WarpRec} not only bridges the gap between academia and industry but also can serve as the architectural backbone for the next generation of sustainable, agent-ready Recommender Systems. Code is available at \url{https://github.com/sisinflab/warprec/}

\end{abstract}





\maketitle

\section{Introduction}
The dual role of Recommender Systems (RS) as scalable industrial tools and rigorous scientific instruments underscores their significance across both commercial and academic sectors.
Hence, the modern RS pipeline is expected to reconcile conflicting imperatives: it must satisfy the massive scalability requirements of industrial applications while adhering to the scientific rigor and ecological responsibility demanded by the research community~\cite{DBLP:journals/access/BarbieratoG24}. Furthermore, with the rise of Agentic AI~\cite{DBLP:conf/www/ZhangHXSMZLW24,DBLP:conf/sigir/WangYZMZ24,DBLP:journals/corr/abs-2503-05659}, recommenders are increasingly tasked with serving as interactive tools for Large Language Models (LLMs) via standardized interfaces such as the Model Context Protocol\footnote{\url{https://modelcontextprotocol.io/}}.
However, the advancement of the field is currently stifled by a fundamental architectural dichotomy between academia and industry. Researchers typically rely on flexible, in-memory libraries (e.g., RecBole~\cite{DBLP:conf/cikm/ZhaoMHLCPLLWTMF21}, Elliot~\cite{DBLP:conf/sigir/AnelliBFMMPDN21}) designed for rapid prototyping. While excellent for algorithmic exploration, these tools rely heavily on eager execution engines (e.g., Pandas) that fail to scale beyond single-node memory limits. Conversely, industrial practitioners utilize distributed frameworks (e.g., NVIDIA Merlin~\cite{DBLP:journals/corr/abs-2210-08803}) optimized for throughput. These systems, while scalable, often lack the flexible evaluation protocols, statistical rigor, and ease of customization required for scientific inquiry.
This divide creates a ``Deployment Chasm'': novel algorithms developed in academia require a complete, costly rewrite to function in production environments~\cite{DBLP:conf/www/ArgyriouGZ20, DBLP:conf/recsys/AmatriainB16}. 

We present \textbf{\textsc{WarpRec}}, a high-performance framework that eliminates these trade-offs. It introduces a backend-agnostic architecture built upon \textit{Narwhals}\footnote{\url{https://narwhals-dev.github.io/narwhals/}}, enabling a ``write-once, run-anywhere'' paradigm. Models defined in \textsc{WarpRec} transition seamlessly from local debugging to distributed training on Ray clusters~\cite{DBLP:conf/osdi/MoritzNWTLLEYPJ18}.
The framework is engineered with extreme modularity; it is composed of five decoupled engines that function as independent, composable units.
This allows researchers to inject custom logic, swap data backends, or utilize the framework's evaluation suite (comprising $40$ metrics and $19$ filtering/splitting strategies) within external pipelines. \textsc{WarpRec} supports $55$ state-of-the-art algorithms, ranging from matrix factorization to the latest graph-based and sequential architectures, all capable of running locally or at cluster scale.
Beyond unifying scale, \textsc{WarpRec} addresses these functional gaps:
\begin{itemize}[leftmargin=*]
    \item \textbf{Green AI \& Sustainability:} It is the first framework to enforce ecological accountability by integrating CodeCarbon\footnote{\url{https://mlco2.github.io/codecarbon/}} for real-time energy tracking and Green AI profiling.
    \item \textbf{Agentic Readiness:} Anticipating the shift toward autonomous systems, \textsc{WarpRec} natively implements the Model Context Protocol server interface. This transforms the recommender from a static predictor into a queryable agent capable of complex reasoning tasks within an LLM loop.
    \item \textbf{Scientific Rigor:} It automates reproducibility, statistical tests and corrections, e.g., Bonferroni~\cite{c3fa9fa7-dd2a-35f2-84a0-d5c07e68dd08} and FDR~\cite{888cd474-50a6-33fd-a789-415b80e67e78}.
\end{itemize}

\begin{figure*}[!t]
\vspace{-1em}
    \centering
    \includegraphics[]{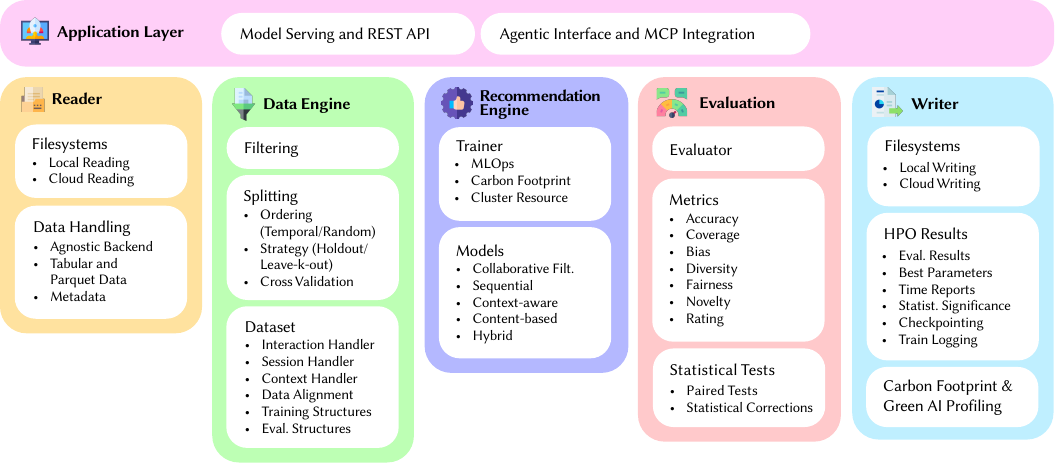}
    \caption{The modular architecture of \textsc{WarpRec}. Five decoupled modules manage the end-to-end recommendation lifecycle, from data ingestion and processing to model training and evaluation. An Application Layer exposes the recommender through a REST API and MCP agentic interface.}
    \label{fig:architecture}
    \vspace{-1em}
\end{figure*}

The remainder of this paper details \textsc{WarpRec}'s architecture, validates its performance through extensive benchmarks against state-of-the-art frameworks, and demonstrates its readiness for the next generation of AI-driven recommendation.

\section{The Fragmented Landscape}
The landscape of recommender frameworks is fractured into distinct silos, none capable of satisfying the demands of modern research.

\vspace{1mm}
\noindent \textbf{The Academic Silo.} 
Academic frameworks have evolved from early tools for classic (MyMediaLite~\cite{DBLP:conf/recsys/GantnerRFS11}, LensKit~\cite{DBLP:conf/cikm/Ekstrand20}, Surprise~\cite{DBLP:journals/jossw/Hug20}) and deep learning models (LibRec~\cite{DBLP:conf/um/GuoZSY15}, Spotlight~\cite{kula2017spotlight}, OpenRec~\cite{DBLP:conf/wsdm/YangBGHE18}) into comprehensive libraries (RecBole \cite{DBLP:conf/cikm/ZhaoMHLCPLLWTMF21}, Cornac \cite{DBLP:journals/jmlr/SalahTL20}) offering hundreds of models and frameworks that prioritize reproducibility~\cite{DBLP:conf/recsys/DacremaCJ19,DBLP:conf/recsys/RendleKZA20} and scientific rigor through automated experimental lifecycles (Elliot~\cite{DBLP:conf/sigir/AnelliBFMMPDN21}, DaisyRec \cite{DBLP:conf/recsys/SunY00Q0G20}).
Despite this progress, they remain confined to a single-node paradigm, that
prevents industrial scaling and necessitates manual infrastructure for distributed hyperparameter optimization (HPO)\cite{DBLP:journals/ijon/YangS20} on engines like Ray~\cite{DBLP:conf/osdi/MoritzNWTLLEYPJ18} or Spark~\cite{DBLP:conf/hotcloud/ZahariaCFSS10} and for using strategies such as ASHA~\cite{DBLP:conf/mlsys/LiJRGBHRT20}. Finally, even rigorous frameworks like Elliot often omit multiple hypothesis testing corrections, leaving them vulnerable to p-hacking~\cite{DBLP:conf/sigir/UrbanoLH19}.

\vspace{1mm}
\noindent \textbf{The Industrial Silo.}
Industrial-grade frameworks such as NVIDIA Merlin~\cite{DBLP:journals/fgcs/PetersonBKRSWAA22}, Apache Spark MLlib~\cite{DBLP:journals/csur/VerbraekenWKKVR20,DBLP:journals/jmlr/MengBYSVLFTAOXX16}, or MS Recommenders~\cite{DBLP:conf/recsys/GrahamMW19} are architected to address the extreme computational demands of production-scale environments. 
These systems overcome the limitations of academic tools by employing distributed dataframes and GPU offloading to facilitate the training of complex architectures on billions of data points~\cite{DBLP:conf/recsys/Cheng0HSCAACCIA16,DBLP:conf/sigir/RongWZZWLLZYGW20}.
However, these tools often prioritize \textit{serving} over \textit{science}, 
lacking fine-grained experimental control. Industrial frameworks usually provide limited data-splitting and evaluation metrics, typically excluding significance testing and beyond-accuracy dimensions 
such as bias, fairness, and diversity~\cite{DBLP:journals/inffus/JinWZZDXP23}. 

\vspace{0.5em}
\noindent \textbf{Green AI.}
The field has pivoted from accuracy-maximizing ``Red AI'' to ``Green AI'', elevating energy efficiency and carbon transparency to first-class metrics alongside traditional performance~\cite{DBLP:journals/cacm/SchwartzDSE20,DBLP:journals/access/BarbieratoG24}.
This is particularly critical in Recommender Systems, where the combination of billion-parameter embedding tables and exhaustive hyperparameter optimization grids leads to extreme energy consumption~\cite{DBLP:conf/recsys/VenteWSB24,DBLP:conf/recsys/SpilloFMMS23}.
Recent works have shown that marginal gains in performance often require exponential increases in carbon emissions~\cite{DBLP:conf/recsys/SpilloFMMS23,DBLP:journals/corr/abs-2410-09359,DBLP:conf/recsys/SpilloFMMS24}.
Despite the critical importance of sustainability, current RS frameworks remain fundamentally energy-blind.
Leading platforms~\cite{DBLP:conf/cikm/ZhaoMHLCPLLWTMF21,DBLP:conf/sigir/AnelliBFMMPDN21,DBLP:journals/jmlr/SalahTL20} provide no native mechanisms, such as CodeCarbon\footnotemark[3] 
 or CarbonTracker~\cite{DBLP:journals/corr/abs-2007-03051}, to quantify carbon emissions or power consumption within the experimental pipeline. 

\vspace{1mm}
\noindent \textbf{Agentic Interoperability in Recommendation.}
Artificial intelligence is shifting from monolithic models to agentic workflows~\cite{DBLP:journals/corr/abs-2503-05659}, where agents call external tools~\cite{DBLP:conf/nips/SchickDDRLHZCS23} and interleave reasoning with actions~\cite{DBLP:conf/iclr/YaoZYDSN023}.
In this paradigm, the recommender becomes a callable tool within an agent's decision-making process~\cite{DBLP:journals/corr/abs-2502-10050}. This new role requires interactive dialogue to iteratively refine results~\cite{DBLP:conf/www/ZhangHXSMZLW24,DBLP:conf/sigir/WangYZMZ24}, yet current frameworks lack the standardized interfaces to enable this. \textsc{WarpRec} addresses this gap by natively implementing the Model Context Protocol, enabling standardized agent-ready interaction.

  \begin{table*}[ht]
\footnotesize
\caption{Comprehensive comparison of \textsc{\framework{}} against frameworks identified in the literature (with year of the last update) across six dimensions: Data, Optimization \& Training, Extensions \& Integration, Rigor \& Reproducibility, Green AI, Metrics, and Statistics.
}
\label{tab:warprec_extended}
\centering
    \renewcommand{\arraystretch}{0.9}
    \setlength{\tabcolsep}{0.50em} 
    \rowcolors{5}{white}{gray!15} 

    \begin{tabular}{l@{\hskip 0.8em}ccc|c|cc|cc|c|cc|cc|cc|cc|cc|cccc|c|ccc}

    \hlineB{2.5}
    \rowcolor{white}
     & \multicolumn{4}{c}{\textbf{Data}} 
     & \multicolumn{4}{c}{\textbf{Optim. \& Train.}} 
     & \multicolumn{3}{c}{\textbf{Ext. \& Int.}}
     & \multicolumn{6}{c}{\textbf{Rigor \& Repro.}}
     & \multicolumn{2}{c}{\textbf{Green AI}} 
     & \multicolumn{5}{c}{\textbf{Metrics}} 
     & \multicolumn{3}{c}{\textbf{Stats}} 
    \\
     \cmidrule(r){2-5}\cmidrule(lr){6-9}\cmidrule(lr){10-12}\cmidrule(lr){13-18}\cmidrule(lr){19-20}\cmidrule(l){21-25}\cmidrule(l){26-28}

    \rowcolor{white}
     & \multicolumn{3}{c}{Backend} & \multicolumn{1}{c}{Mgmt}
     & \multicolumn{2}{c}{Env} & \multicolumn{2}{c}{Tuning} &  \multicolumn{1}{c}{Ext.} & \multicolumn{2}{c}{Int.} & \multicolumn{2}{c}{Tests} & \multicolumn{2}{c}{Corr.}
     & \multicolumn{2}{c}{Repro.} 
     & \multicolumn{2}{c}{Report.} 
     & \multicolumn{4}{c}{Bey. Acc.} & \multicolumn{1}{c}{Comp.} & \multicolumn{3}{c}{Catalog}
    \\
     \cmidrule(r){2-4}\cmidrule(lr){5-5}\cmidrule(lr){6-7}
     \cmidrule(lr){8-9}\cmidrule(lr){10-10}\cmidrule(lr){11-12}
     \cmidrule(lr){13-14}\cmidrule(lr){15-16}\cmidrule(lr){17-18}
     \cmidrule(lr){19-20}\cmidrule(lr){21-24}\cmidrule(lr){25-25} 
     \cmidrule(lr){26-28}

    \rowcolor{white}
     & \rotatebox{90}{Eager (Pandas)} 
     & \rotatebox{90}{Fast (Polars)} 
     & \rotatebox{90}{Distrib. (Spark)} 
     & \rotatebox{90}{Cloud I/O}
     
     & \rotatebox{90}{Multi-GPU (DDP)}
     & \rotatebox{90}{Cluster (Ray)}
     & \rotatebox{90}{Search Alg.} 
     & \rotatebox{90}{Pruning (ASHA)}

     & \rotatebox{90}{Callbacks}
     & \rotatebox{90}{Agentic AI}
     & \rotatebox{90}{REST API}
     
     & \rotatebox{90}{Paired (T/Wilc)}
     & \rotatebox{90}{Indep (Kru/Whi)}
     & \rotatebox{90}{FWER (Bonf)}
     & \rotatebox{90}{FDR (Ben-Hoch)}
     & \rotatebox{90}{Seed}
     & \rotatebox{90}{Checkpoint}
     & \rotatebox{90}{Carbon Track}
     & \rotatebox{90}{Energy Report}
     & \rotatebox{90}{Bias}
     & \rotatebox{90}{Fairness}
     & \rotatebox{90}{Nov/Div} 
     & \rotatebox{90}{Multi-Objective}
     & \rotatebox{90}{GPU Accel.} 
    
     & \rotatebox{90}{\# Models}
     & \rotatebox{90}{\# Split/Filtering}
     & \rotatebox{90}{\# Metrics}
     \\ 
    \toprule

    \hlineB{2.0}
    \myrowcolour
     \multicolumn{28}{c}{\textbf{Classic}} \\
     \toprule
     \toprule
     
    LibRec (2022)~\cite{DBLP:conf/um/GuoZSY15} & & & & & & & \cmark & & & & & & & & & \cmark & & & & 0& 0& 0& 0& & 55 & 10 & 11\\
    MyMediaLite (2020)~\cite{DBLP:conf/recsys/GantnerRFS11} & & & & &  & & \cmark & & & & & & & & & \cmark & & & & 0& 0& 0& 0& & 18 & 6 & 8 \\
    Surprise (2024)~\cite{DBLP:journals/jossw/Hug20} &  & & & \cmark & & & \cmark & &  & & & & & & & \cmark & & & & 0& 0& 0& 0& & 7 & 6 & 4 \\

    \toprule
    \hlineB{2.0}
    \myrowcolour
     \multicolumn{28}{c}{\textbf{Research}} \\
     \toprule
     \toprule
    LensKit (2026)~\cite{DBLP:conf/cikm/Ekstrand20} & \cmark & \cmark & & & \cmark& \cmark& \cmark & & & & & & & & & \cmark & \cmark& & & 2 & 0 & 3 & 0 & & 12 & 5 & 11 \\
    RecBole (2025)~\cite{DBLP:conf/cikm/ZhaoMHLCPLLWTMF21} & \cmark & & & & \cmark & & \cmark & \cmark & & & & & & & & \cmark & \cmark & & & 2 & 0 &  3 & 0 & \cmark & 159 & 8 & 16 \\ 
    DaisyRec (2024)~\cite{DBLP:conf/recsys/SunY00Q0G20} & \cmark & & & & & & \cmark & &  & & & & & & & \cmark & & & & 1& 0& 2& 0 & & 13 & 7 & 11 \\
    
    Spotlight (2020)~\cite{kula2017spotlight}  & & & & & & & \cmark & & & & & & & & & & & & & 0& 0& 0& 0 & & 2 &  2 & 4 \\
Elliot (2023)~\cite{DBLP:conf/sigir/AnelliBFMMPDN21} & \cmark & & & &  & & \cmark &  &  & & & \cmark & & & & \cmark & & & & 5 & 9 & 6 & 0 & & 77 & 13 & 38 \\ 
    
    \toprule
    \hlineB{2.0}
    \myrowcolour
     \multicolumn{28}{c}{\textbf{Specialized (Diversity, Knowledge, Evaluation)}} \\
     \toprule
     \toprule
Cornac (2025)~\cite{DBLP:journals/jmlr/SalahTL20} & \cmark & & & & & & \cmark & & &  & & & & & & \cmark & & & & 0& 0& 0 & 0 & & 58 & 4 & 12 \\ 
    FuxiCTR (2025)~\cite{DBLP:journals/corr/abs-2009-05794} & \cmark & \cmark & & & \cmark& & \cmark & & & & & & & & & \cmark & \cmark & & & 0& 0&0 & 0 &\cmark & 51 & 2 & 4 \\
    ClayRS (2024)~\cite{DBLP:journals/is/LopsPMSS23} & \cmark & & & & & & \cmark & & & & & \cmark & \cmark & \cmark & \cmark& \cmark& & & &1& 1& 3& 0 & & 17 & 3 & 15 \\
    Informfully (2026)~\cite{DBLP:conf/recsys/HeitzLIB25} &  \cmark & & & & & & \cmark & &  & & \cmark & & &  & & \cmark & & & & 0 & 0& 10 &0 & & 10 & 5 & 22 \\
    RecList (2023)~\cite{DBLP:conf/www/ChiaTBHK22} & \cmark & & & \cmark & & & & & \cmark & & & & & & & \cmark & & & & 1 & 3 & 1 & 0 && - & - & 15\\

    \toprule
    \hlineB{2.0}
    \myrowcolour
     \multicolumn{28}{c}{\textbf{Industry Scale}} \\
     \toprule
     \toprule

    MS Recommenders (2026)~\cite{DBLP:conf/recsys/GrahamMW19} & \cmark & & \cmark & \cmark & \cmark & & \cmark & & \cmark  & & \cmark & & & & & \cmark & \cmark & & &0 &1 &4 & 0 && 15 & 3 & 15 \\
    Merlin (2024)~\cite{DBLP:journals/corr/abs-2210-08803} & \cmark & \cmark & \cmark & \cmark & \cmark & \cmark & \cmark & \cmark & \cmark  & & \cmark & & & &  & \cmark & \cmark & & & 0& 0& 0& 0 &\cmark & 8 & 8 &  5\\

    \midrule
    \textsc{\framework} 
    & \textbf{\cmark} & \textbf{\cmark} & \textbf{\cmark} & \textbf{\cmark} & \textbf{\cmark} & \textbf{\cmark} & \textbf{\cmark} & \textbf{\cmark} & \textbf{\cmark} & \textbf{\cmark} & \textbf{\cmark} & \textbf{\cmark} & \textbf{\cmark} 
    & \textbf{\cmark} & \textbf{\cmark} & \textbf{\cmark} & \textbf{\cmark} & \textbf{\cmark} & \textbf{\cmark} & 5 & 10 & 6 
    & 2 &\textbf{\cmark} 
    & 55 & 19 & 40  \\ 

    \bottomrule
    \end{tabular}
\end{table*}

\section{\textsc{WarpRec}}
The architecture of \textsc{WarpRec}, illustrated in Figure~\ref{fig:architecture}, is designed 
with 
principles of modularity and separation of concerns, 
ensuring each phase of the recommendation lifecycle is independent and extensible. \Cref{tab:warprec_extended} compares its features 
against
established frameworks. 

\subsection{Pipelines and Callbacks}
To streamline the research lifecycle, \textsc{WarpRec} abstracts complex workflows into three standardized execution pipelines:
\begin{itemize}[leftmargin=*]
    \item \textbf{Training Pipeline}, which orchestrates the full end-to-end process, automating hyperparameter optimization (HPO);
    \item \textbf{Design Pipeline}, optimized for rapid prototyping and architectural validation without HPO overhead;
    \item \textbf{Evaluation Pipeline}, dedicated to post-hoc analysis and inference using pre-trained checkpoints.
\end{itemize}
These workflows are controlled via declarative configuration files, avoiding boilerplate orchestration code. 
This high degree of modularity is specifically intended for industrial applications and advanced custom workflows. Rather than being restricted to the predefined pipelines, practitioners can extract and integrate individual modules into existing production ecosystems.
The framework further integrates an event-driven \textbf{Callback} system allowing custom hooks at specific stages, facilitating the injection of custom logic and enabling complex experiments without modifying the core pipeline.

\subsection{Reader Module}
\label{sec:reader}

The \textit{Reader Module} is designed to efficiently ingest user-item interactions and metadata, abstracting the complexity of data retrieval. The module leverages Narwhals\footnotemark[2] as a backend-agnostic compatibility layer. Instead of enforcing a specific data structure, this design allows to abstract the data loading process, uniforming the data access interface without incurring costly conversion overheads.
Furthermore, the module decouples logical data access from physical storage. It supports seamless ingestion from both local filesystems and cloud-based object storage. This architectural choice ensures that the framework remains environment-agnostic, enabling researchers to transition effortlessly from local prototyping to large-scale cloud experimentation without modifying their ingestion pipelines.

\subsection{Data Engine}
The \textit{Data Engine} module is responsible for transforming raw transactions into refined, model-ready structures. To ensure experimental rigor and flexibility, this module is composed of three specialized components: (i) Filtering, (ii) Splitting, and (iii) Dataset Management.

\vspace{0.2em}
\noindent \textbf{Filtering.}
The \textit{Filtering} component includes 13 distinct strategies, organized into three functional families~\cite{DBLP:conf/sigir/AnelliBFMMPDN21}: 
(i) \textit{filter-by-rating} strategies, which mitigate noise by pruning interactions based on global or distributional value thresholds; 
(ii) \textit{$k$-core} decomposition, which iteratively removes users and items with fewer than $k$ connections; and 
(iii) \textit{cold-user/item} heuristics, which target the cold-start problem by discarding entities that fail to meet minimum interaction constraints.

\vspace{0.2em}
\noindent \textbf{Splitter.}
To prevent data leakage and ensure rigorous evaluation~\cite{DBLP:reference/sp/GunawardanaSY22}, the \textit{Splitter} component partitions the filtered dataset using 6 distinct strategies. 
\textsc{WarpRec} supports both \textit{random} and \textit{temporal} configurations for \textit{Hold-Out} and \textit{Leave-$k$-Out} protocols, \textit{Fixed Timestamp} splitting and $k$-fold \textit{Cross-Validation}. 
The module enforces system-wide determinism by anchoring all stochastic partitioning to global random seeds, guaranteeing experimental reproducibility.

\vspace{1mm}
\noindent \textbf{Dataset.} This component orchestrates the transformation of raw partitions into high-performance internal representations. It unifies \textit{Interaction}, \textit{Session}, and \textit{Context Management} to standardize sparse user-item signals, variable-length behavioral sequences, and auxiliary metadata. The \textit{Dataset} enforces strict \textit{Data Alignment} by mapping external identifiers to contiguous internal indices, ensuring referential integrity before materializing optimized \textit{Training and Evaluation Structures} designed for high-throughput ingestion.

\subsection{Recommendation Engine}
This module governs the lifecycle and training of diverse model architectures, enabling training, hyperparameter search, and resource management. It includes two components: (i) Models and (ii) Trainer.


\vspace{0.2em}
\noindent \textbf{Models.}
This component encapsulates recommendation algorithms as decoupled entities, independent of data ingestion and evaluation protocols. 
In its current release, the framework provides a robust repository of 55 built-in algorithms spanning 6 fundamental classes~\cite{DBLP:reference/sp/2022rsh}: (i) \textit{Unpersonalized}, (ii) \textit{Content-Based}, (iii) \textit{Collaborative Filtering} (CF), (iv) \textit{Context-Aware}, (v) \textit{Sequential}, and (vi) \textit{Hybrid} models.
Within the CF family, the framework covers several dominant paradigms, including: Autoencoders (e.g., EASE$^R$~\cite{DBLP:conf/www/Steck19}, MultiVAE~\cite{DBLP:conf/www/LiangKHJ18}); Graph-based architectures (e.g., LightGCN~\cite{DBLP:conf/sigir/0001DWLZ020}, DGCF~\cite{DBLP:conf/sigir/WangJZ0XC20}); K-Nearest Neighbors (e.g., UserKNN~\cite{DBLP:conf/cscw/ResnickISBR94}, ItemKNN~\cite{DBLP:conf/www/SarwarKKR01}); Latent Factor models (e.g., BPRMF~\cite{DBLP:journals/corr/abs-1205-2618}, SLIM~\cite{DBLP:conf/icdm/NingK11}); and Neural recommenders (e.g., NeuMF~\cite{DBLP:conf/www/HeLZNHC17}, ConvNCF~\cite{DBLP:journals/tois/DuHYTQC19}). To handle heterogeneous metadata, \textsc{WarpRec} incorporates robust factorization-based baselines--such as FM~\cite{DBLP:conf/icdm/Rendle10}, AFM~\cite{DBLP:conf/ijcai/XiaoY0ZWC17}, and NFM~\cite{DBLP:conf/sigir/0001C17}--alongside deep architectures designed for high-dimensional sparse and dense features, including Wide\&Deep~\cite{DBLP:conf/recsys/Cheng0HSCAACCIA16}, DeepFM~\cite{DBLP:conf/ijcai/GuoTYLH17}, and xDeepFM~\cite{DBLP:conf/kdd/LianZZCXS18}. The Sequential suite is equally comprehensive, encompassing CNN-based (Caser~\cite{DBLP:conf/wsdm/TangW18}), Markov-Chain (FOSSIL~\cite{DBLP:conf/icdm/HeM16}), RNN (GRU4Rec~\cite{DBLP:journals/corr/HidasiKBT15}), and models based on Transformer architecture (BERT4Rec~\cite{DBLP:conf/cikm/SunLWPLOJ19}).
To enable modern research, \textsc{WarpRec} integrates a wide range of state-of-the-art models, including LightCCF~\cite{DBLP:conf/sigir/ZhangZZ0025}, EGCF~\cite{DBLP:journals/tois/ZhangZSS25}, ESIGCF~\cite{DBLP:journals/eaai/YangHCLQ25}, MixRec~\cite{DBLP:conf/www/ZhangZ25}, LightGCN++\cite{DBLP:conf/recsys/LeeKS24}, gSASRec\cite{DBLP:conf/ijcai/PetrovM24}, and LinRec~\cite{DBLP:journals/corr/abs-2411-01537}. 

\vspace{0.2em}
\noindent \textbf{Trainer.}
This component serves as the framework's core execution engine, orchestrating model optimization and state persistence via automated checkpointing to enable seamless experiment resumption. Designed to bridge the gap between academic research and production MLOps, it supports distributed computing across multi-GPU clusters while incorporating comprehensive experiment tracking hooks for rigorous monitoring.

\vspace{0.2em}
\noindent \textit{Execution Environments and Distributed Training.}
\textsc{WarpRec} ensures strict code portability from local prototyping to industrial deployment through seamless vertical and horizontal scaling. The framework supports execution from single-node to multi-GPU configurations, leveraging Ray for multi-node orchestration~\cite{DBLP:conf/osdi/MoritzNWTLLEYPJ18}. This integration enables elastic scaling across cloud infrastructures, optimizing resource allocation and reducing computational costs.

\vspace{0.2em}
\noindent \textit{Hyperparameter Tuning.}
To automate hyperparameter discovery, the module integrates robust HPO strategies ranging from Grid and Random search to Bayesian and bandit-based optimization (e.g., \textit{HyperOpt}~\cite{DBLP:conf/icml/BergstraYC13}, \textit{Optuna}~\cite{DBLP:conf/kdd/AkibaSYOK19}, \textit{BoHB}~\cite{DBLP:conf/icml/FalknerKH18}) over heterogeneous search spaces. Computational efficiency is maximized via a dual-layer stopping mechanism: system-level pruning using the \textit{ASHA} scheduler~\cite{DBLP:conf/mlsys/LiJRGBHRT20} and model-level early stopping based on convergence metrics.

\vspace{0.2em}
\noindent \textit{Dashboarding.}
To facilitate real-time observability, the framework integrates TensorBoard, Weights \& Biases, and MLflow~\cite{DBLP:journals/debu/ZahariaCD0HKMNO18}. 
In
line with Green AI principles~\cite{DBLP:conf/mlsys/WuRGAAMCBHBGGOM22}, CodeCarbon\footnotemark[3] was adopted to automatically quantify energy consumption and carbon emissions.

\subsection{Evaluation}
To guarantee scientific reproducibility, the \textit{Evaluation} module provides a suite of multidimensional metrics while supporting statistical hypothesis testing and error correction.

\vspace{0.5em}
\noindent \textbf{Metrics.}
\textsc{WarpRec} integrates a comprehensive suite of 40 metrics organized into distinct functional families: \textit{Accuracy}~\cite{DBLP:conf/kdd/ZhouZSFZMYJLG18,Schroder201178}, \textit{Rating}, \textit{Coverage}, \textit{Novelty}~\cite{DBLP:conf/recsys/VargasC11}, \textit{Diversity}~\cite{DBLP:conf/sigir/ZhaiCL03}, \textit{Bias}~\cite{DBLP:conf/flairs/AbdollahpouriBM19,DBLP:conf/recsys/AbdollahpouriBM17,DBLP:journals/pvldb/YinCLYC12,DBLP:conf/sigir/ZhuWC20,DBLP:conf/recsys/TsintzouPT19}, and \textit{Fairness}~\cite{deldjoo2020flexible,DBLP:conf/cikm/ZhuHC18}.
As highlighted in Table~\ref{tab:warprec_extended}, \textsc{WarpRec} distinguishes itself as the sole framework to natively support \textit{Multi-objective} metrics~\cite{DBLP:conf/emo/ZitzlerBT06, DBLP:journals/isci/NoiaRTS17}. This capability advances the state-of-the-art by enabling model selection through the optimization of competing goals (e.g., balancing accuracy and popularity bias).
To ensure scalability, the metric computation is fully GPU-accelerated, drastically reducing the latency of large-scale experimental loops.

\vspace{0.5em}

\noindent \textbf{Statistical Hypothesis Testing.}
To guarantee methodological rigor~\cite{DBLP:conf/sigir/Sakai16, DBLP:journals/sigir/Fuhr17}, \textsc{WarpRec} integrates a comprehensive hypothesis testing suite into the evaluation pipeline. The framework automates significance testing, supporting both paired comparisons (e.g., Student's t-test~\cite{The-Probable-Error-of-a-Mean}, Wilcoxon signed-rank~\cite{Wilcoxon1945}) and independent-group analyses (e.g., Mann-Whitney U~\cite{Mann-Whitney_U}). \textsc{WarpRec} mitigates the Multiple Comparison Problem~\cite{DBLP:journals/tois/Carterette12} through Bonferroni~\cite{bonferroni1936teoria} and FDR~\cite{FDR-1995} corrections for Type I errors, ensuring that performance gains are statistically robust.

\subsection{Writer Module}
The \textit{Writer Module} ensures reproducibility and streamlined monitoring via a storage-agnostic interface that persists artifacts to local or cloud backends. Beyond standard performance tables and statistical significance tests, \textsc{WarpRec} automatically serializes granular per-user metrics, optimized hyperparameters, and trained model weights. Furthermore, the framework logs recommendation lists, execution metadata, execution times, and carbon-emission estimates, supporting both rigorous experimentation and sustainable AI research.

\subsection{Application Layer}
Finally, \textsc{WarpRec} bridges the gap between rapid experimentation and production deployment through a strictly modular architecture. Unlike monolithic frameworks, \textsc{WarpRec} decouples the modeling core from the training infrastructure, allowing trained models to be isolated as standalone artifacts with zero additional engineering effort. This design empowers a versatile model serving layer where specific recommenders can be instantly exposed via: (i) RESTful APIs for high-throughput, real-time inference in standard microservices; and (ii) an MCP server, enabling Large Language Models and autonomous agents to dynamically query the recommender as a tool.





\section{Performance and Multi-faceted Analysis}

We show that \textsc{WarpRec} bridges the gap between academic rigor and industrial scalability by answering the following research questions:

\begin{itemize}[leftmargin=*]
    \item \textbf{RQ1: Scalability and Performance.} How do \textsc{WarpRec}'s architectural optimizations impact performance when scaling from academic prototyping to massive industrial datasets?
    
    \item \textbf{RQ2: Green AI Analysis.}  Can \textsc{WarpRec} effectively quantify  the carbon footprint in large-scale recommendation workflows?
    
    \item \textbf{RQ3: Agentic AI.} Does \textsc{WarpRec} provide the modularity and abstraction to serve as a backend for autonomous agentic systems?
\end{itemize}

\begin{table}[t]
\footnotesize
\caption{Statistical summary of the experimental datasets.}
\label{tab:datasets_stats}
\centering
    \renewcommand{\arraystretch}{0.9}
    \setlength{\tabcolsep}{1.08em} 
    \rowcolors{2}{gray!15}{white}

    \begin{tabular}{lrrrr}
    \toprule
    \textbf{Dataset} & \textbf{\# Users} & \textbf{\# Items} & \textbf{\# Interactions} & \textbf{Sparsity} \\
    \midrule
    
    MovieLens-1M & 6,040 & 3,883 & 1,000,209 & 95.7353\% \\
    MovieLens-32M & 200,948 & 87,585 & 32,000,204 & 99.8182\% \\
    NetflixPrize-100M & 480,189 & 17,770 & 100,480,507 & 98.8281\% \\

    \bottomrule
    \end{tabular}\vspace{-0.4em}
\end{table}
\begin{table}[t]
\centering
\footnotesize
\renewcommand{\arraystretch}{0.85}
\setlength{\tabcolsep}{0.39em} 
\caption{Computational resource requirements (CPUs / GPUs / RAM in GB) by dataset and execution mode. \textit{P}: Parallel; \textit{S}: Serial.}
\label{tab:resources}
\begin{tabular}{llccccc}
\toprule
\textbf{Data} & \textbf{Mode} & \textbf{EASE$^R$} & \textbf{ItemKNN} & \textbf{LightGCN} & \textbf{NeuMF} & \textbf{SASRec} \\
\midrule
\multirow{2}{*}{MovieLens-1M} & S & 16/1/64 & -- & 16/1/64 & 16/1/64 & 16/1/64 \\
& P & -- & -- & 32/1/64 & 32/1/64 & 32/2/64 \\
\midrule
\multirow{2}{*}{MovieLens-32M} & S & 16/1/256 & -- & 16/1/128 & 16/1/128 & 16/1/128 \\
& P & -- & -- & 32/4/128 & 32/4/128 & 32/4/128 \\
\midrule
\multirow{2}{*}{NetflixPrize-100M} & S & 16/1/192 & 16/1/192 & 16/1/192 & 16/1/192 & 16/1/192 \\
& P & -- & -- & 32/4/192 & 32/4/192 & 32/4/192 \\
\bottomrule
\end{tabular}%
\end{table}
\begin{table}
\caption{End-to-end performance and scalability benchmark. WarpRec is compared against leading frameworks on datasets of increasing scale; $\dagger$ denotes its serial execution without parallelization. Best results are reported in bold, second best are \underline{underlined}.}
\label{tab:performance_comparison}
\centering
\footnotesize
    \renewcommand{\arraystretch}{0.85}
    \setlength{\tabcolsep}{0.4em} 

    \begin{tabular}{llrrrrrrr}
    \bottomrule
        \myrowcolour
    \multicolumn{9}{c}{\textbf{NetflixPrize-100M (Medium Scale)}} \\ 
    \toprule

     & \textbf{Fram.} & \textbf{C} & \textbf{Prep.} & \textbf{Train} & \textbf{Eval} & \textbf{HPO} & \textbf{Total} & \textbf{nDCG} \\        
    
    \midrule


    \multirow{6}{*}{\rotatebox{90}{\textbf{EASE$^R$}}} 
    & Cornac & 1 & 7m 58s & \textbf{2m 35s} & 29m 45s & \underline{3h 14m} & \underline{3h 22m} & 0.3143 \\
    & Elliot & 1 & 2m 39s & \underline{2m 54s} & 5h 17m & \multicolumn{3}{c}{\cellcolor{gray!15}\textit{Time Limit Exceeded (4/6)}} \\
    & RecBole & 1 & 11m 17s & 26m 29s & \textbf{20m 21s} & 4h 41m & 5h 49m & 0.3283 \\
    & DaisyRec & 1 & \underline{1m 35s} & 1h 17m & \multicolumn{4}{c}{\cellcolor{gray!15}\textit{Out of Memory - RAM}} \\
    & \framework{}$^\dagger$ & 1 & \textbf{30.25s} & 3m 9s & \underline{21m 13s} & \textbf{2h 26m} & \textbf{2h 26m} & 0.3035 \\
    \midrule

    \multirow{6}{*}{\rotatebox{90}{\textbf{NeuMF}}} 
    & Cornac & 1 & 7m 33s & 19m 55s & 16m 24s & 3h 38m & 3h 45m & 0.1267 \\
    & Elliot & 1 & \multicolumn{6}{c}{\cellcolor{gray!15}\textit{Time Limit Exceeded (0/6)}}\\
    & RecBole & 1 & 11m 13s & 1h 23m & 36m 57s & 11h 59m & 13h 6m & 0.1080 \\
    & DaisyRec & 1 & 1m 36s & 2h 39m & \multicolumn{4}{c}{\cellcolor{gray!15}\textit{Out of Memory - VRAM}} \\
    & MS Rec. & 1 & 46m 13s & \multicolumn{5}{c}{\cellcolor{gray!15}\textit{Time Limit Exceeded (0/6)}} \\
    & \framework{}$^\dagger$ & 1 & \underline{29.95s} & \textbf{9m 13s} & \underline{2m 30s} & \underline{1h 10m} & \underline{1h 10m} & 0.1323 \\
    & \framework{} & 6 & \textbf{21.87s} & \underline{19m 34s} & \textbf{2m 29s} & \textbf{28m 0s} & \textbf{28m 22s} & 0.1323 \\
    \midrule

    \multirow{6}{*}{\rotatebox{90}{\textbf{LightGCN}}} 
    & Cornac & 1 & 7m 46s & 10h 4m & 12m 26s & \multicolumn{3}{c}{\cellcolor{gray!15}\textit{Time Limit Exceeded (2/6)}} \\
    & Elliot & 1 & 2m 33s & 15h 53m & 42.13s & \multicolumn{3}{c}{\cellcolor{gray!15}\textit{Time Limit Exceeded (1/6)}} \\
    & RecBole & 1 & 11m 33s & \textbf{1h 27m} & 4m 19s & \textbf{9h 7m} & \textbf{10h 17m} & 0.0247 \\
    & DaisyRec & 1 & 1m 55s & 14h 21m  & 1m 7s & \multicolumn{3}{c}{\cellcolor{gray!15}\textit{Time Limit Exceeded (1/6)}} \\
    & MS Rec. & 1 & 4m 36s & \multicolumn{5}{c}{\cellcolor{gray!15}\textit{Out of Memory - VRAM}} \\
    & \framework{}$^\dagger$ & 1 & \underline{29.53s} & \underline{4h 53m} & \textbf{39.15s} & \multicolumn{3}{c}{\cellcolor{gray!15}\textit{Time Limit Exceeded (2/6)}} \\
    & \framework{} & 6 & \textbf{26.38s} & 14h 35m & \underline{40.81s} & \underline{22h 4m} & \underline{22h 5m} & 0.1779 \\
    \midrule

    \multirow{3}{*}{\rotatebox{90}{\textbf{SASRec}}} 
    & RecBole & 1 & 42m 37s & 2h 37m & 2m 59s & 15h 59m & 20h 15m & 0.0515 \\
    & \framework{}$^\dagger$ & 1 & \textbf{21.19s} & \textbf{36m 32s} & \underline{51.85s} & \underline{3h 44m} & \underline{3h 44m} & 0.1204 \\
    & \framework{} & 6 & \underline{24.86s} & \underline{1h 12m} & \textbf{51.68s} & \textbf{1h 56m} & \textbf{1h 56m} & 0.1204 \\


    \bottomrule
    \myrowcolour
    \multicolumn{9}{c}{\textbf{MovieLens-32M (Small Scale)}} \\
    \toprule

     & \textbf{Fram.} & \textbf{C} & \textbf{Prep.} & \textbf{Train} & \textbf{Eval} & \textbf{HPO} & \textbf{Total} & \textbf{nDCG} \\       

        \midrule

    \multirow{6}{*}{\rotatebox{90}{\textbf{EASE$^R$}}}
    & Cornac & 1 & 1m 58s & \textbf{26m 46s} &  \multicolumn{4}{c}{\cellcolor{gray!15}\textit{Out of Memory - RAM}} \\
    & Elliot & 1 & 50.71s & 34m 8s & 10h 12m & \multicolumn{3}{c}{\cellcolor{gray!15}\textit{Time Limit Exceeded (2/6)}} \\
    & RecBole & 1 & 3m 19s & 51m 0s & \textbf{17m 28s} & \underline{6h 51m} & \underline{7h 10m} & 0.3609 \\
    & DaisyRec & 1 & \underline{24.93s} & 3h 9m & \multicolumn{4}{c}{\cellcolor{gray!15}\textit{Out of Memory - RAM}} \\
    & \framework{}$^\dagger$ & 1 & \textbf{9.343s} & \underline{29m 2s} & \underline{32m 14s} & \textbf{6h 7m} & \textbf{6h 7m} & 0.3399 \\
    \midrule

    \multirow{6}{*}{\rotatebox{90}{\textbf{NeuMF}}} 
    & Cornac & 1 & 2m 9s & \underline{27m 13s} & 21m 59s & 4h 55m & 4h 57m & 0.1264 \\
    & Elliot & 1 & 16h 9m & 2h 26m & 26m 19s & \multicolumn{3}{c}{\cellcolor{gray!15}\textit{Time Limit Exceeded (3/6)}} \\
    & RecBole & 1 & 3m 29s & 10h 51m & 1h 3m & \multicolumn{3}{c}{\cellcolor{gray!15}\textit{Time Limit Exceeded (2/6)}} \\
    & DaisyRec & 1 & 25.20s & 1h 44m & \multicolumn{4}{c}{\cellcolor{gray!15}\textit{Out of Memory - VRAM}} \\
    & MS Rec. & 1 & 2h 6m & \multicolumn{5}{c}{\cellcolor{gray!15}\textit{Time Limit Exceeded (0/6)}} \\
    & \framework{}$^\dagger$ & 1 & \underline{9.70s} & \textbf{13m 0s} & \textbf{4m 55s} & \underline{1h 47m} & \underline{1h 47m} & 0.1456 \\
    & \framework{} & 6 & \textbf{9.03s} & 58m 21s & \underline{4m 57s} & \textbf{1h 34m} & \textbf{1h 34m} & 0.1456 \\
    \midrule
    
    \multirow{6}{*}{\rotatebox{90}{\textbf{LightGCN}}} 
    & Cornac & 1 & 1m 58s & 5h 59m & 20m 32s & \multicolumn{3}{c}{\cellcolor{gray!15}\textit{Time Limit Exceeded (3/6)}} \\
    & Elliot & 1 & 49.73s & 7h 37m & \textbf{15.45s} & \multicolumn{3}{c}{\cellcolor{gray!15}\textit{Time Limit Exceeded (2/6)}} \\
    & RecBole & 1 & 3m 11s & \textbf{29m 53s} & 1m 51s & \textbf{3h 10m} & \textbf{3h 29m} & 0.0237 \\
    & DaisyRec & 1 & 26.07s & 6h 21m & 1m 17s & \multicolumn{3}{c}{\cellcolor{gray!15}\textit{Time Limit Exceeded (3/6)}} \\
    & MS Rec. & 1 & 1m 16s & \multicolumn{5}{c}{\cellcolor{gray!15}\textit{Out of Memory - VRAM}} \\
    & \framework{}$^\dagger$ & 1 & \underline{10.34s} & \underline{3h 47m} & 39.54s & 22h 46m & 22h 46m & 0.2061 \\
    & \framework{} & 6 & \textbf{9.14s} & 7h 16m & \underline{38.94s} & \underline{10h 7m} & \underline{10h 7m} & 0.2061 \\
    \midrule
    
    \multirow{3}{*}{\rotatebox{90}{\textbf{SASRec}}} 
    & RecBole & 1 & 12m 48s & 4h 11m & 1m 6s & \multicolumn{3}{c}{\cellcolor{gray!15}\textit{Time Limit Exceeded (5/6)}} \\
    & \framework{}$^\dagger$ & 1 & \underline{8.18s} & \textbf{1h 0m} & \underline{40.24s} & \underline{6h 4m} & \underline{6h 4m} & 0.0889 \\
    & \framework{} & 6 & \textbf{7.91s} & \underline{1h 45m} & \textbf{39.94s} & \textbf{2h 36m} & \textbf{2h 36m} & 0.0889 \\

    \bottomrule
    \myrowcolour
    \multicolumn{9}{c}{\textbf{MovieLens-1M (Tiny Scale)}} \\
    \toprule

     & \textbf{Fram.} & \textbf{C} & \textbf{Prep.} & \textbf{Train} & \textbf{Eval} & \textbf{HPO} & \textbf{Total} & \textbf{nDCG} \\       

        \midrule

    \multirow{6}{*}{\rotatebox{90}{\textbf{EASE$^R$}}} 
    & Cornac & 1 & 2.97s & \textbf{1.31s} & 7.21s & \underline{51.15s} & \underline{54.12s} & 0.2736 \\
    & Elliot & 1 & 2.14s & \underline{1.44s} & 13.60s & 1m 30s & 1m 32s & 0.2920 \\   
    & RecBole & 1 & 5.39s & 7.38s & \underline{3.76s} & 1m 7s & 1m 39s & 0.3221 \\
    & DaisyRec & 1 & \underline{0.59s} & 46.85s & 5.25s & 5m 12s & 5m 13s & 0.0888 \\
    & \framework{}$^\dagger$ & 1 & \textbf{0.33s} & 2.32s & \textbf{2.77s} & \textbf{30.52s} & \textbf{30.85s} & 0.2850 \\
    \midrule

    \multirow{6}{*}{\rotatebox{90}{\textbf{NeuMF}}} 
    & Cornac & 1 & 2.90s & 47.40s & 7.51s & 5m 29s & 5m 32s & 0.1927 \\
    & Elliot & 1 & 14.70s & 1m 0s & 2.13s & 6m 14s & 6m 29s & 0.2209 \\
    & RecBole & 1 & 6.41s & 2m 17s & 10.27s & 14m 47s & 15m 25s & 0.1664 \\
    & DaisyRec & 1 & \textbf{0.55s} & 1m 10s & \textbf{0.66s} & 7m 6s & 7m 7s & 0.0030 \\
    & MS Rec. & 1 & 25.97s & 3m 11s & 2m 43s & 35m 25s & 35m 51s & 0.2630 \\
    & \framework{}$^\dagger$ & 1 & \underline{0.94s} & \textbf{18.93s} & 1.26s & \underline{2m 1s} & \underline{2m 2s} & 0.1943 \\
    & \framework{} & 6 & 0.97s & \underline{32.70s} & \underline{1.25s} & \textbf{52.48s} & \textbf{53.45s} & 0.1943 \\
    \midrule
    
    \multirow{6}{*}{\rotatebox{90}{\textbf{LightGCN}}} 
    & Cornac & 1 & 2.93s & 2m 36s & 4.67s & 16m 6s & 16m 9s & 0.2482 \\
    & Elliot & 1 & 1.69s & 1m 43s & \textbf{0.36s} & 10m 24s & 10m 26s & 0.2269 \\
    & RecBole & 1 & 5.29s & \underline{44.21s} & 2.81s & 4m 42s & 5m 13s & 0.0216 \\
    & DaisyRec & 1 & \textbf{0.51s} & 1m 16s & \underline{0.46s} & 7m 40s & 7m 41s & 0.0512 \\
    & MS Rec. & 1 & 1.40s & 1m 0s & 1.74s & 7m 53s & 7m 54s & 0.1860 \\
    & \framework{}$^\dagger$ & 1 & \underline{0.95s} & \textbf{22.26s} & 0.87s & \underline{2m 19s} & \underline{2m 20s} & 0.2122 \\
    & \framework{} & 6 & 1.02s & 1m 26s & 0.82s & \textbf{1m 41s} & \textbf{1m 42s} & 0.2122 \\
    \midrule
    
    \multirow{3}{*}{\rotatebox{90}{\textbf{SASRec}}} 
    & RecBole & 1 & 23.16s & 7m 3s & 1.72s & 42m 34s & 44m 53s & 0.0689 \\
    & \framework{}$^\dagger$ & 1 & \underline{0.99s} & \textbf{1m 55s} & \textbf{1.01s} & \underline{11m 36s} & \underline{11m 37s} & 0.0649 \\
    & \framework{} & 6 & \textbf{0.95s} & \underline{4m 49s} & \underline{1.12s} & \textbf{6m 26s} & \textbf{6m 27s} & 0.0649 \\
    
    \bottomrule
    \end{tabular}
\end{table}

\subsection{Experimental Setup}
\label{sec:experimental}

To rigorously evaluate the scalability and efficiency of \textsc{WarpRec}, we designed a comprehensive experimental protocol spanning a diverse suite of recommendation frameworks, datasets of varying magnitude, and strict reproducibility standards.

\vspace{0.2em}
\noindent\textbf{Baselines.}
We consider frameworks that provide experimentation pipelines out-of-the-box, eliminating error-prone custom implementations (e.g., NVIDIA Merlin). Accordingly, we compare against 5 leading frameworks, such as Cornac~\cite{DBLP:journals/jmlr/SalahTL20}, DaisyRec~\cite{DBLP:journals/pami/SunFYQLYOZ23}, Elliot~\cite{DBLP:conf/sigir/AnelliBFMMPDN21}, and RecBole~\cite{DBLP:conf/cikm/ZhaoMHLCPLLWTMF21} for academic benchmarking, and Microsoft Recommenders~\cite{DBLP:conf/recsys/GrahamMW19} for enterprise-scale deployment.

\vspace{0.2em}
\noindent\textbf{Models.}
We benchmark five representative algorithms that span different learning paradigms, including EASE$^R$~\cite{DBLP:conf/www/Steck19}, NeuMF~\cite{DBLP:conf/www/HeLZNHC17}, LightGCN~\cite{DBLP:conf/sigir/0001DWLZ020}, SASRec~\cite{DBLP:conf/icdm/KangM18} and ItemKNN~\cite{DBLP:conf/www/SarwarKKR01} (included as a reference baseline for environmental impact comparisons within our analysis).

\vspace{0.2em}
\noindent\textbf{Datasets.}
Our benchmarking spans three scaling datasets, from academic prototyping to industrial deployment size:  MovieLens-1M, MovieLens-32M\footnote{Both available at \url{https://grouplens.org/datasets/movielens/}}, 
and NetflixPrize-100M~\cite{DBLP:journals/sigkdd/BellK07}. The dataset statistics are summarized in Table~\ref{tab:datasets_stats}. We limited the dataset scale to permit competitor execution success, as \textsc{WarpRec}'s capacity extends further.

\vspace{0.2em}
\noindent\textbf{Splitting Strategy.}
We employ 90-10 random holdout for EASE$^R$, LightGCN, NeuMF, ItemKNN and temporal holdout for SASRec.

\vspace{0.2em}
\noindent\textbf{Optimization Protocol.}
Models are tuned via exhaustive grid search of $6$ configurations per model from established literature~\cite{DBLP:conf/um/AnelliBNJP22, DBLP:journals/access/BetelloPSTBTS25} (details on GitHub\footnote{\url{https://github.com/sisinflab/warprec-benchmark-2026/}}). We train for 10 epochs on MovieLens and 2 on Netflix-100M. Best models are selected via validation nDCG@10, using a 8,192 batch size and a 24h timeout per trial. Given the memory requirement of EASE$^R$ it is only considered for serial execution.



\vspace{0.2em}
\noindent\textbf{Hardware and Resources.}
Experiments were conducted on 16-core CPUs and 64GB NVIDIA A100 GPUs. We compare \textit{serial} (sequential) and \textit{parallel} (concurrent via Ray) execution modes to evaluate scalability. Resource allocations are detailed in Table~\ref{tab:resources}.

\vspace{0.2em}
\noindent\textbf{Evaluation Metrics.}
We evaluate \textsc{WarpRec}'s efficiency, considering $C$ concurrent trials, by measuring:
(i) Preprocessing time, including ingestion and splitting; (ii) Training and (iii) Evaluation times, measured as the average per-trial optimization and validation; (iv) HPO duration, the total wall-clock time for hyperparameter exploration\footnote{\textsc{WarpRec} can perform per-epoch evaluation, but we excluded it since others do not.}; and the total time aggregates preprocessing and HPO.
Finally, we report nDCG@10 with full ranking evaluation to ensure that speed-ups do not compromise algorithmic fidelity.

\vspace{0.2em}
\noindent\textbf{Reproducibility and Carbon Footprint.}
Experiments are tracked via Weights \& Biases using fixed seed, while CodeCarbon monitors energy and emissions. All code and configurations are on GitHub\footnotemark[5]. 







\begin{table}[t]
\caption{Green AI Profiling of \textsc{WarpRec} on NetflixPrize-100M, captured via CodeCarbon. Best results are reported in bold.} 
\label{tab:green_ai_detailed}
\centering
\footnotesize
    \renewcommand{\arraystretch}{0.9}
    \setlength{\tabcolsep}{0.64em} 
    \rowcolors{2}{gray!15}{white}

    \begin{tabular}{lrrrrr}
    \toprule
    \textbf{Metric} & \textbf{ItemKNN} & \textbf{EASE$^R$} & \textbf{NeuMF} & \textbf{LightGCN} & \textbf{SASRec} \\
    \midrule
    Emissions & \textbf{0.0002} & 0.0005 & 0.0004 & 0.0095 & 0.0012 \\
    Emissions Rate & \textbf{2.96$e^{-7}$} & 3.09$e^{-7}$ & 4.97$e^{-7}$ & 4.00$e^{-7}$ & 5.33$e^{-7}$ \\
    CPU Power & \textbf{137.8144} & 155.3458 & 219.4410 & 154.1137 & 152.7022 \\
    GPU Power & 75.9096 & \textbf{71.7731} & 177.2569 & 157.3857 & 278.5554 \\
    CPU Energy & \textbf{0.0248} & 0.0638 & 0.0530 & 1.0235 & 0.0967 \\
    GPU Energy & \textbf{0.0133} & 0.0292 & 0.0424 & 1.0373 & 0.1847 \\
    RAM Energy & \textbf{0.0095} & 0.0220 & 0.0129 & 0.3584 & 0.0345 \\
    Energy Consumed & \textbf{0.0476} & 0.1150 & 0.1083 & 2.4192 & 0.3159 \\
    Peak RAM Usage & 74.8316 & 67.5962 & 99.3873 & \textbf{53.6099} & 86.1017 \\

    \bottomrule
    \end{tabular}
\begin{tablenotes}
\footnotesize
\item Emissions: kg CO$_2$eq; Emissions Rate: kg CO$_2$eq/h; Power: W; Energy: kWh; RAM: GB.
\end{tablenotes}
\end{table}

\subsection{RQ1: Scalability and Performance}
\label{sec:scalability_and_performance}


To address RQ1, Table~\ref{tab:performance_comparison} details the efficiency results for increasing scale datasets of \textsc{WarpRec} against academic and industrial competitor frameworks.
The failures to complete the pipeline can be due to either memory exhaustion (Out-of-Memory errors in both RAM and VRAM) or exceeding the execution time limits. They are highlighted in the table using a gray box starting at the specific stage where execution was interrupted. In the case of \textit{Time Limit Exceeded} errors, the notation $(x/6)$ indicates the $x$ number of hyperparameter configurations successfully completed before the timeout. 
The experimental results reveal significant disparities in how various frameworks handle increasing data scales. On the smallest dataset (MovieLens-1M), all evaluated frameworks completed the full recommendation pipeline, confirming their suitability for small-scale academic prototyping. However, as the data scale increases to MovieLens-32M and NetflixPrize-100M, critical nuances regarding architectural robustness emerge. Several established frameworks exhibit systemic limitations when tasked with medium-to-large-scale experiments. These frameworks frequently fail to complete the pipeline. In contrast, \textsc{WarpRec} consistently executes the end-to-end workflow across all benchmarks, positioning it as a robust infrastructure for both high-stakes research and industrial deployment. Notably, LightGCN is the most computationally intensive model across all frameworks. Except for RecBole and \textsc{WarpRec}, all other frameworks failed to process the NetflixPrize-100M dataset for this model. 
Due to the tight constraints of our experimental setup, RecBole's LightGCN implementation struggled to reach convergence within the time limit.
Notably, algorithms such as EASE$^R$ (frequently favored in industrial applications for their balance of simplicity and
performance~\cite{DBLP:conf/recsys/SpisakBHPT23, DBLP:conf/recsys/VancuraAKK22}) require the inversion of a dense Gram matrix, which poses a significant challenge to data engines. \textsc{WarpRec} outperforms all competitors across all dataset scales for this model.

\begin{figure}[t]
\centering
\resizebox{\columnwidth}{!}{
    \begin{tcolorbox}[colback=gray!15, colframe=gray!95, arc=2mm, boxrule=1pt,left=4pt, right=4pt, top=4pt, bottom=4pt]
    \footnotesize{
        \textbf{User:} `` I've recently watched Pulp Fiction, Forrest Gump, and Full Metal Jacket. Do you have any suggestions for what I should watch next? ''
        
        
        \begin{tcolorbox}[colback=gray!5, colframe=gray!20, boxrule=0.5pt, arc=1.5mm, left=3pt, right=3pt, top=3pt, bottom=3pt]
            \textit{\textbf{Agent (Internal Thought):} User requests recommendations based on viewing history. Calling $\rightarrow$ \texttt{WarpRec SASRec Tool}...}
            
            \vspace{3pt}
            \textit{\textbf{Tool Call:} \texttt{WarpRec\_SASRec.recommend (item\_sequence=[`Pulp Fiction', `Forrest Gump', `Full Metal Jacket'], top\_k=3)}}
            
            \vspace{3pt}
            \textit{\textbf{WarpRec Tool Output:} [1. `Star Wars: Episode IV - A New Hope', 2. `The Godfather', 3. `Schindler's List']}
        \end{tcolorbox}
        
        \textbf{Agent:} `` Here are some movies you could like:
        \textbf{The Godfather} would be an excellent choice given your appreciation for character-driven storytelling and grittier narratives. For something epic and immersive, the original \textbf{Star Wars IV} captures the iconic '70s era you seem to enjoy. \textbf{Schindler's List} shares the historical depth and emotional weight of your recent watches. Do you prefer something more action-adventure oriented, or another powerful drama? ''
    }
    \end{tcolorbox}
}
\caption{Sequential Recommendation via \textsc{WarpRec} MCP Interface. The AI Agent leverages a SASRec model trained on MovieLens-32M.
}
\label{fig:agentic_example}
\vspace{-1em}
\end{figure}

\subsection{RQ2: Green AI Analysis}
Leveraging \textsc{WarpRec}'s native integration with CodeCarbon, we profiled the environmental impact of training recommendation models on the NetflixPrize-100M dataset (Table~\ref{tab:green_ai_detailed}). A cross-analysis with end-to-end performance (Table~\ref{tab:performance_comparison}) reveals that total energy consumption is driven more by training duration than by instantaneous power draw. For instance, while SASRec exhibits the highest peak GPU power (278.6 W), its relatively rapid convergence results in a moderate total energy footprint (0.32 kWh). In contrast, LightGCN, despite a lower average power draw (157.4 W), requires a significantly more extensive training phase to converge, resulting in a massive aggregate consumption of 2.42 kWh and the highest carbon emissions (0.0095 kg CO$_2$eq). Conversely, shallow architectures demonstrate a superior balance between efficiency and effectiveness; specifically, EASE$^R$ consumes approximately $95\%$ less energy (0.115 kWh) than the deep graph-based baselines, demonstrating that state-of-the-art recommendation performance can be achieved with minimal environmental cost.

\subsection{RQ3: Interoperability with Agentic Pipelines}
To address RQ3, we assess the capability of \textsc{WarpRec}'s  \textit{Application Layer}, which abstracts away the complexities of serving, ensuring that trained models are instantly available for inference without modification.
As illustrated in Figure~\ref{fig:agentic_example}, the system performs real-time inference using a SASRec model trained on the MovieLens-32M dataset (achieving an nDCG@10 of $0.9257$ with $100$ negative samples). Thanks to its sequential nature, the model does not require a pre-trained user embedding to process the input history.
After receiving the response from the \textsc{WarpRec} tool, the agent generates a natural language explanation identifying latent semantic connections to enrich the raw recommendation list. The recommendations are presented highlighting ``character-centered storytelling'' and ``more raw storytelling,'' thus combining the algorithmic result and agentic reasoning. This shows how \textsc{WarpRec}'s MCP interface bridges the gap between recommendation mechanisms and agentic workflows.

\section{Conclusion}
\textsc{WarpRec} represents a paradigm shift in Recommender Systems engineering, resolving the long-standing dichotomy between academic prototyping and industrial recommendation through a backend-agnostic architecture for reproducible, scalable, and responsible research.
Our analysis demonstrates that \textsc{WarpRec} achieves industrial-grade throughput while retaining the flexibility of lightweight academic libraries. It shows that a single codebase can easily transition from local exploration to distributed training on Ray clusters.
By operationalizing Green AI principles and enforcing statistical rigor by default, \textsc{WarpRec} ensures that future innovations are scientifically valid and environmentally responsible. Finally, \textsc{WarpRec} transforms the recommender from a static ranking engine into a dynamic, queryable partner for Agentic AI.
\bibliographystyle{ACM-Reference-Format}
\bibliography{bibliography}










\end{document}